\documentclass{article}

\usepackage{makeidx}  

\usepackage{graphicx}
\usepackage{cite}

\usepackage{color}
\usepackage{amsmath}
\usepackage{epsfig}
\usepackage{url}

\usepackage{bm}

\usepackage{amssymb}
\usepackage{amsfonts}

\usepackage{graphicx}
\usepackage{cite}

\usepackage{color}
\usepackage{amsmath}
\usepackage{epsfig}
\usepackage{url}

\usepackage{bm}

\usepackage{amssymb}
\usepackage{amsfonts}

\newcommand{\bfx}{{\bf x}}

\newcommand{\bfr}{{\bf r}}

\newcommand{\R}{{\bf R}}
\newcommand{\A}{{\bf A}}
\newcommand{\B}{{\bf B}}

\newcommand{\I}{{\bf I}}

\newcommand{\X}{{\bf X}}

\title{Learning LiNGAM based on data with more variables than observations}
\author{Shohei Shimizu\thanks{The Institute of Scientific and Industrial Research (ISIR), Osaka University, Mihogaoka 8-1, Ibaraki, Osaka 567-0047, Japan. Email: sshimizu@ar.sanken.osaka-u.ac.jp}}
\date{}

\begin{document}

\maketitle

\begin{abstract}
A very important topic in systems biology is developing statistical methods that automatically find causal relations in gene regulatory networks with no prior knowledge of causal connectivity. 
Many methods have been developed for time series data. 
However, discovery methods based on steady-state data are often necessary and preferable since obtaining time series data can be more expensive and/or infeasible for many biological systems. 
A conventional approach is causal Bayesian networks. 
However, estimation of Bayesian networks is ill-posed. In many cases it cannot uniquely identify the underlying causal network and only gives a large class of equivalent causal networks that cannot be distinguished between based on the data distribution. 
We propose a new discovery algorithm for uniquely identifying the underlying causal network of genes.
To the best of our knowledge, the proposed method is the first algorithm for learning gene networks based on a fully identifiable causal model called LiNGAM.
We here compare our algorithm with competing algorithms using artificially-generated data, although it is definitely better to test it based on real microarray gene expression data. 
\end{abstract}

\section{Introduction}\label{intro}
An important topic in bioinformatics is developing computational methods to discover gene regulatory causal networks from static expression data \cite{Cantone09Cell,diBernardo05Nature,Peer11Cell}. 
Based on the estimated networks, one can compute intervention effects, {\it i.e.}, causal effects, which enable predicting to what extent the expression level of a gene changes when that of another gene is externally changed \cite{Pearl00book}. 
One can rank causal relations between genes according to the existense and/or strengths of causal effects. 
Such a ranking can be used as a priority list to efficiently conduct future interventional experiments and obtain solid evidence \cite{Maathuis10NatureMethods}. 

Many estimation techniques have been proposed for time series data \cite{Bansal06Bioinformatics,Rangel04Bioinformatics,Opgen06WCSB,Perrin03Bioinformatics}. 
Those techniques use temporal information to estimate the underlying gene network structure.
 However, it is not very feasible to obtain time series data for many biological systems. 
In fact, many data that have been analyzed as time series data are not really longitudinal due to destructive sampling \cite{Oates12AOAS}. 
Then, discovery methods based on steady-state data could be better suited for such non-longitudinal `time series' data. 

A conventional approach for estimating gene causal networks based on steady state data is Bayesian networks \cite{Friedman00JofComputationalBiology,Peer01Bioinformatics}. 
However, Bayesian networks suffers from the identifiability problem. In the framework of Bayesian networks, many networks with different structures give the same conditional independences between variables or genes, and in many cases one cannot uniquely estimate the underlying causal network without any prior knowledge on the structure \cite{Pearl00book,Spirtes93book}. Thus, causal effects often are not uniquely estimated as well. 

In this paper, we propose an attractive alternative approach that enables uniquely estimating a causal gene network. 
We first model the causal network of genes using a non-Gaussian causal model called LiNGAM \cite{Shimizu06JMLR}. LiNGAM is a fully identifiable causal model \cite{Shimizu06JMLR} unlike conventional Bayesian networks  and has recently attracted much attention in machine learning \cite{Spirtes10Tribute,Moneta11NIPS09}. 
Then, we present a new algorithm for estimating LiNGAM based on data with more variables than observations. 
Finally, we test our approach using artificially-generated data. 

\section{Methods}\label{sec:method}
We first define our model in Section~\ref{sec:model} and then propose a new algorithm to estimate the model based on data with more variables than observations in Section~\ref{sec:est}. 

\subsection{Model}\label{sec:model}
Let us denote by $x_i$ the expression level of gene $i$ $(i=1, \cdots, p)$. 
We model causal relations of gene expression levels (variables) $x_i$ $(i=1, \cdots, p)$ using a linear non-Gaussian causal model called LiNGAM \cite{Shimizu06JMLR}: 
\begin{eqnarray}
x_i &=& \sum_{k(j)<k(i)} b_{ij}x_j + e_i, \label{eq:model}
\end{eqnarray}
where $k(i)$ is such a causal ordering of genes that they graphically form a directed acyclic graph (DAG). This means that no later gene directly or indirectly regulates any earlier gene, that is, has a directed path on any earlier gene. The $e_i$ are the external influences or noises and $b_{ij}$ are connection strengths of gene $j$ on gene $i$. 
The zero/non-zero pattern of $b_{ij}$ corresponds to the absence/existence pattern of directed edges. 
External influences $e_i$ follow non-Gaussian continuous distributions with non-zero variances and are mutually independent. 

The assumption that external influences $e_i$ are non-Gaussian enables unique identification of a causal ordering $k(i)$ and connection strengths $b_{ij}$ without using any background knowledge on the structure \cite{Shimizu06JMLR}. 
This feature is a big advantage \cite{Shimizu06JMLR,Spirtes10Tribute,Moneta11NIPS09} over conventional Bayesian network approaches based on conditional independences and/or Gaussianity \cite{Spirtes91,Chic02JMLR}. 
Though the Gaussian approximation has been a common approach \cite{Hyva01book}, real-world data could be considered more or less non-Gaussian. 
In fact, non-Gaussian data appear in many applications including bioinformatics \cite{Frigyesi06BMCBioinformatics,Sogawa11NN}. 

We note that each $b_{ij}$ ($i \neq j$) represents the direct causal effect of $x_j$ on $x_i$ and each $a_{ij}$ ($i \neq j$), the $(i,j)$-th element of the matrix $\A$$=$$(\I-\B)^{-1}$, the total causal effect of $x_j$ on $x_i$ \cite{Pearl00book,Hoyer07IJAR}. 
Based on total causal effects $a_{ij}$, one can predict to what extent the expression level of gene $i$ changes when that of gene $j$ is externally changed. 
Further, based on direct causal effects $b_{ij}$, one can predict to what extent the expression level of gene $i$ changes when that of gene $j$ is externally changed while those of all other genes than gene $i$ and gene $j$ are held fixed.

Rigorously speaking, the linearity assumption would be more or less violated in real-world gene regulatory causal networks. 
Nonlinear approaches \cite{Imoto03JBCB,Hoyer09NIPS,Zhang09UAI} might be better to model causal relations of genes. 
However, in general, linear methods can often give better results when it is more important to find quantitative relations since nonlinear methods usually require very large sample sizes \cite{Peer11Cell}. 


\subsection{Estimation of the model}\label{sec:est}
Now, the problem of causal discovery is to estimate a causal ordering $k(i)$ and connection strengths $b_{ij}$ based on data $\bfx$ only.
Several estimation methods for LiNGAM \cite{Shimizu06JMLR,Shimizu11JMLR} have been proposed that do not require to specify the distributions of the non-Gaussian external influences $e_i$. 
However, they are not applicable for such cases with more variables than observations that are typical in gene expression data analysis. 
Thus, we extend an LiNGAM estimation algorithm \cite{Shimizu11JMLR} to cases with more variables than observations. 

In \cite{Shimizu11JMLR}, a direct method was proposed to estimate a causal ordering $k(i)$. 
This leads to more algorithmically reliable results since it is not necessary to resort to iterative search in the parameter space. 
First, it estimates an exogenous variable. 
An exogenous variable is a variable with no parents and can be at the top of a causal ordering. 
One can estimate such a variable that is exogenous by finding a variable that minimizes a non-parametric independence measure called KGV \cite{Bach02JMLR} between the variable and its regression residuals \cite{Shimizu11JMLR}. 
Once an exogenous variable is found, 
then one subtract the effect of the exogenous variable from the other variables using linear least squares regression.  
One can find all the causal orders by iterating this \cite{Shimizu11JMLR}. 
 Once a causal ordering $k(i)$  is estimated, one can prune or set to zero redundant connection strengths among $b_{ij}$ by repeatedly applying a sparse regularization method called adaptive lasso \cite{Zou06JASA} on each variable and its potential parents.
The adaptive lasso \cite{Zou06JASA} is a weighted version of a regularization technique for variable selection lasso \cite{Tibshirani96JRSSB} and assumes the same data generating process as LiNGAM: 
\begin{equation}
x_i = \sum_{k(j)<k(i)} b_{ij}x_j + e_i.\nonumber \label{eq:lasso}
\end{equation}
The adaptive lasso assumes that the set of such potential parent variables $x_j$ that $k(j)$$<$$k(i)$ is known, while LiNGAM estimates the set of such variables. 
The adaptive Lasso penalizes connection strengths $b_{ij}$ in $L_1$ penalty by minimizing the objective function defined as: 
\begin{eqnarray}
\left\| x_i-\sum_{k(j)<k(i)} b_{ij}x_j \right\|^2 + \lambda \sum_{k(j)<k(i)} w_{ij}|b_{ij}|,\nonumber
\end{eqnarray}
where $\lambda$ is a regularization parameter and $w_{ij}$ is a weight for $b_{ij}$. 
In \cite{Zou06JASA}, it was suggested to use the inverse of the absolute value of the ordinary least squares regression estimate or the ridge regression estimate of $b_{ij}$ as $w_{ij}$. 
The adaptive lasso asymptotically selects the right set of such variables $x_j$ that $b_{ij}$ is not zero, where $k(j)$$<$$k(i)$. 

To apply the direct method \cite{Shimizu11JMLR} on data with more variables than observations, we make three modifications. 
First, in cases with more variables than observations, the sample covariance matrix of variables is singular. Thus, we use ridge regression instead of linear least squares regression to compute regression residuals. 
Second, non-parametric independence measures, e.g., KGV \cite{Bach02JMLR} and HSIC \cite{Gretton05ALT}, require large sample sizes and much computational time. They could give a similar performance as simple nonlinear correlation measures for small sample sizes \cite{Entner10AMBN}. 
Therefore, we use a nonlinear correlation \cite{Shimizu09UAI,Entner10AMBN} to evaluate independence. We first evaluate pairwise independence between a variable and each of the residuals and then take the sum of the pairwise independence measures over the residuals. 
Third, before applying adaptive lasso, we reduce the dimension of data to at most $n$$-$$1$ using a dimension reduction method called iterative sure independence screening (ISIS) \cite{Fan08JRSSB} so that the dimension is smaller than the sample size. 
This would be reasonable in estimation of gene networks since they are commonly assumed to be sparse. 
ISIS selects explanatory variables that have the first $n$$/$$\log(n)$ largest correlation coefficients with the explained variable in absolute value and apply lasso on the selected variables. It repeats this procedure on the residual of the explained variable over the selected explanatory variables until a desired number of variables, here $n$$-$$1$ variables, are selected. 
We then apply lasso on the selected variables by ISIS and finally adaptive lasso on the variables still surviving. 
The combination of ISIS, lasso and adaptive lasso was suggested by \cite{Fan08JRSSB} to accurately identify non-zero coefficients when the dimension is larger than the sample size. 
We select the regularization parameter for each method using BIC based on Gaussianity \cite{Zhang10JASA,Zou09AS}. 
Though our model assumes non-Gaussianity,  we optimistically assume that the effect of misspecification of the model distribution might not be so serious since the lasso and adaptive lasso estimation only involves means and covariances of observed variables. 
Moreover, we can estimate total causal effects $a_{ij}$ by applying ISIS, lasso and adaptive lasso of gene $i$ on the set of gene $j$ and the parents of gene $j$ according to the back-door criterion \cite{Pearl95Biometrika}. 



We now present our algorithm to estimate the LiNGAM in Equation~(\ref{eq:model}) from data with more variables (or genes) than observations: 

\noindent
  \rule{\columnwidth}{0.5mm}
       { \sffamily
{\bf Input}: Data matrix $\X$
	 \begin{enumerate}
	 \item Given a $p$-dimensional random vector $\bfx$, a set of its variable subscripts $U$ and a $p \times n$ data matrix of the random vector as $\X$, initialize an ordered list of variables $K:=\emptyset$ and $m:=1$. 
	 \item Repeat until $p$$-$$1$  subscripts are appended to $K$:
	 \begin{enumerate}
	 \item \label{step:2a} Denote by $\bfx_K$ a vector that collects all the variables in $K$. 
	 Perform ridge regression of $x_j$ on $\bfx_K$ with the ridge parameter $\tau$ and compute the residual $\widetilde{x}_j$ for all $j \in U \backslash  K$. 
	  Compute the matrix $\widetilde{\X}$ that collects the values of $\widetilde{x}_j$ ($j \in U \backslash  K$) from the data matrix $\X$. 
	 \item  Perform ridge regression of $x_i$ on $[x_j, \bfx_K^T]^T$  with the ridge parameter $\tau$ for all $i \in U \backslash  K$ ($i \neq j$) and compute the residual vectors $\bfr^{(j)}$$=$$[r_i^{(j)}]$ and the residual data matrix $\R^{(j)}$ from the matrix $\X$ for all $j \in U \backslash  K$. 
	 \item Find a variable $\widetilde{x}_m$ that is most independent of the residuals $r_i^{(j)}$:  	 
	\begin{eqnarray}
	\widetilde{x}_m = \arg\min_{j \in U \backslash  K} \sum_{i\in U\backslash K, i \neq j} |{\rm corr} \{g(\widetilde{x}_j), r_{i}^{(j)}\}| \nonumber \\
	+  |{\rm corr} \{\widetilde{x}_j, g(r_{i}^{(j)})\}|, \label{eq:T}
	\end{eqnarray}	
	where $g(\cdot)$ is $\tanh(\cdot)$. 
	\item Append $m$ to the end of $K$. 		
	 \end{enumerate}
	 \item Append the remaining variable subscript to the end of $K$.
	 \item \label{step:reg} Estimate connection strengths or direct causal effects $b_{ij}$ by doing adaptive lasso of $x_i$ on all the variables with earlier causal orders than $x_i$ based on the data matrix $\X$ for all $i \in U$. In case of more parent candidates than observations, the dimension is reduced to at most $n-1$ by ISIS and lasso before applying adaptive lasso. 
	 The weights $w_{ij}$ for adaptive lasso are estimated by the inverses of the absolute values of ridge regression coefficients with the ridge parameter $\tau$.
	 \item Estimate total causal effects $a_{ij}$ by doing adaptive lasso of $x_i$ on the set of $x_j$ and its parent variables based on the data matrix $\X$ for all $i,j \in U (i\neq j)$. 
	 In case of more parent candidates than observations, the dimension is reduced to at most $n-1$ by ISIS and lasso before applying adaptive lasso.  
	 The weights $w_{ij}$ for adaptive lasso are estimated by the inverses of the absolute values of ridge regression coefficients with the ridge parameter $\tau$.	 
	 \end{enumerate}
	 	 {\bf Output}: A causal network given by the zero/non-zero pattern of $\B$, direct causal effects $b_{ij}$ and total causal effects $a_{ij}$ between observed variables $x_i$ and $x_j$ ($i,j=1,\cdots,p, i\neq j$). \\
       } \vspace{-4mm}
\noindent \rule{\columnwidth}{0.5mm}
\vspace{2mm}

\section{Experiments on artificial data}\label{exp:artificial}
As a sanity check of our method, we performed an experiment with synthetic data. 
The experiment consisted of 1001 trials. 
In each trial, we generated a dataset with dimension $p=100$ and sample size $n=30$ and applied our estimation method on the data. 
For comparison,  we also tested three methods: 1) random guessing, 2) lasso \cite{Tibshirani96JRSSB} and 3) elastic net \cite{Zou05JRSSB}. 
Regarding lasso and elastic net, we applied lasso or elastic net on every variable taking the other variables as its explanatory variables and considered that explanatory variables with non-zero regression coefficients have non-zero direct causal effects and non-zero total causal effects on the explained variable as in \cite{Maathuis10NatureMethods} and \cite{Shimamura11PlosOne}. Note that the lasso- and elastic net-based approaches do not give a DAG network. 
We used the coordinate descent algorithm in \cite{Friedman10JSS} of Matlab statistics toolbox to perform lasso and elastic net. Regarding elastic net, the weights of the lasso penalty and ridge penalty were set to be equal. 
All the regularization parameters were selected using BIC based on Gaussianity. 
The ridge parameter $\tau$ for ridge regression was set to 0.01. 
Regarding random guessing, we first generated a random ordering of observed variables and then randomly created as many non-zero direct causal effects and non-zero total causal effects as LiNGAM found while keeping the network being acyclic.  
 
Each dataset was created as follows: 
\begin{enumerate}
\item We constructed the $p \times p$ connection strength matrix with all zeros and replaced every element in the lower-triangular part by independent realizations of Bernoulli random variables with success probability $s$ similarly to \cite{Kalisch07JMLR}. The probability $s$ determines the sparseness of the model. 
The expected number of adjacent variables of each observed variable is given by $s(p-1)$. 
We randomly set the sparseness $s$ so that the expected number of adjacent variables was 2 or 5. 
\item We replaced each non-zero entry in the connection strength matrix by a value randomly chosen from the interval $[-1.5, -0.5]$ $\cup$ $[0.5, 1.5]$ and selected variances of the external influences from the interval $[1, 3]$. 
The resulting matrix was used as the data-generating connection strength matrix $\B$. 
\item \label{step:2} We generated data with sample size $n$ by independently drawing the external influence variables $e_i$. We randomly selected the distribution of each $e_i$ from a multimodal asymmetric mixture of two Gaussians, a multimodal symmetric mixture of two Gaussians and a Laplace distributions used in \cite{Bach02JMLR}. 
Then, we generated constants following the Gaussian distribution $N(0,4)$ and added them to the external influence variables as their means. 
\item The values of the observed variables $x_i$ were generated using the connection strength matrix $\B$ and external influences $e_i$. 
Finally, we permuted the variables according to a random ordering. 
\end{enumerate}
 
\begin{figure}[!tb]
\begin{center}
  \includegraphics[width=0.8\textwidth]{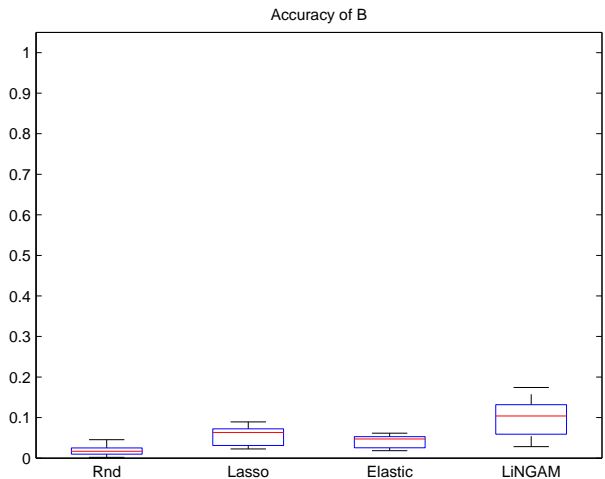}
\end{center}
\vspace{-4mm}
\caption{Box plot of accuracies of non-zero direct causal effects $b_{ij}$.}
\label{fig:accuracy_B}       
\end{figure}

\begin{figure}[!tb]
\begin{center}
  \includegraphics[width=0.8\textwidth]{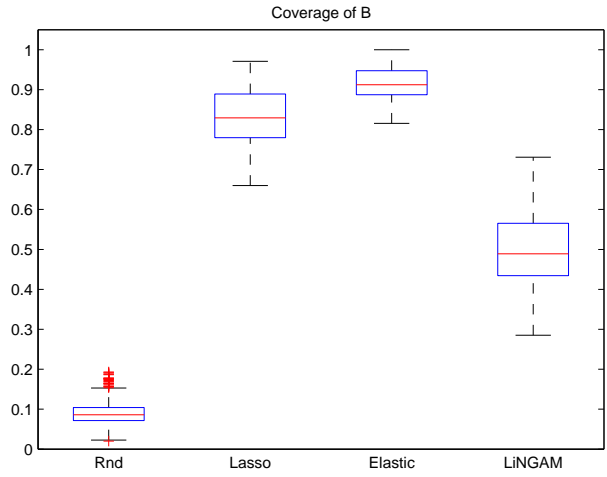}
\end{center}
\vspace{-4mm}
\caption{Box plot of coverages of non-zero direct causal effects $b_{ij}$.}
\label{fig:coverage_B}       
\end{figure}

\begin{figure}[!tb]
\begin{center}
  \includegraphics[width=0.8\textwidth]{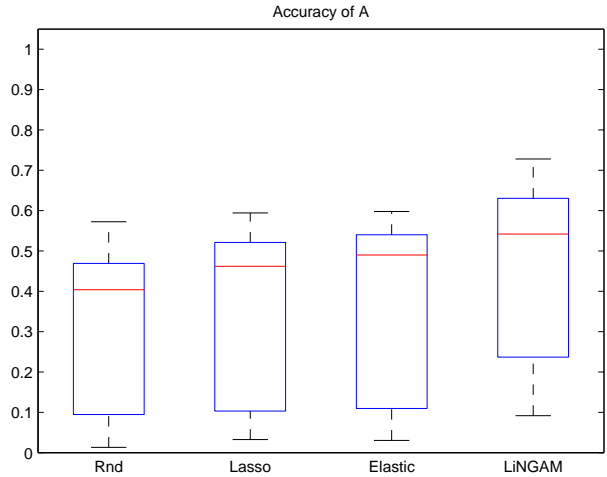}
\end{center}
\vspace{-4mm}
\caption{Box plot of accuracies of non-zero total causal effects $a_{ij}$.}
\label{fig:accuracy_A}       
\end{figure}

\begin{figure}[!tb]
\begin{center}
  \includegraphics[width=0.8\textwidth]{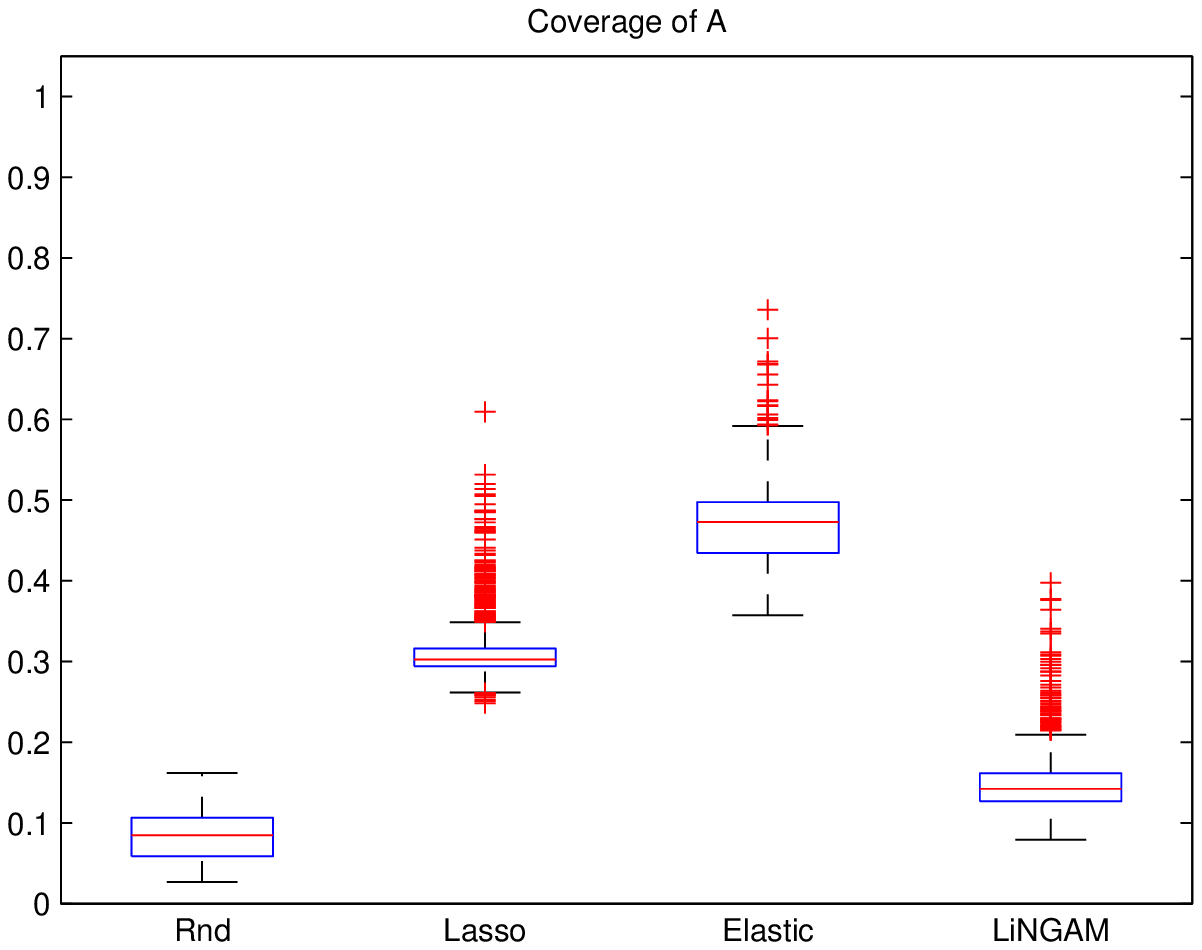}
\end{center}
\vspace{-4mm}
\caption{Box plot of coverages of non-zero total causal effects $a_{ij}$.}
\label{fig:coverage_A}       
\end{figure}

For every dataset, we computed the percentage of actually non-zero direct causal effects (or existing directed edges) in estimated ones, that is, the accuracy (or the true discovery rate) and the percentage of correctly estimated non-zero direct causal effects in actually non-zero ones, that is, the coverage. 
Fig.~\ref{fig:accuracy_B} and Fig.~\ref{fig:coverage_B} show box plots of the accuracies and coverages of non-zero direct causal effects in $\B$. 
The median accuracies of non-zero direct causal effects for the four methods, random guessing, lasso, elastic net, and LiNGAM, were 
0.017,  0.063, 0.047, and 0.104, respectively and the median coverages were 0.086, 0.829, 0.912, and 0.489, respectively. 
Fig.~\ref{fig:accuracy_A} and Fig.~\ref{fig:coverage_A} show box plots of the accuracies and coverages of non-zero total causal effects in $\A$($=(\I-\B)^{-1}$). 
The median accuracies of non-zero total causal effects for random guessing, lasso, elastic net, and LiNGAM were 0.404, 0.462, 0.490, and 0.54, respectively and the median coverages were 0.085, 0.303, 0.473, and  0.142, respectively. 
The median computational times for the lasso, elastic net and LiNGAM were 60.13, 48.36, and 728 seconds on a standard PC. 

In summary, our method gave better accuracies than the other methods. This implied that our method is more suitable for prioritizing future experiments.

\section{Conclusions}\label{sec:conc}
We presented a new algorithm for estimating LiNGAM based on data with more variables than observations. 
Future works would include i) empirical comparison of our method and related algorithms on microarray gene expression datasets;  
ii) extensions of our method to cases with latent confounding variables \cite{Hoyer07IJAR} and cyclic relations \cite{Lacerda08UAI}.

\section*{Acknowledgement}
We would like to thank Satoshi Hara for discussions. 


\bibliography{shimizu12a}
\bibliographystyle{unsrt}

\end{document}